\documentclass[letterpaper, 10 pt, journal, twoside]{IEEEtran}

\usepackage{cite}

\ifCLASSINFOpdf
  \usepackage[pdftex]{graphicx}
\else
  \usepackage[dvips]{graphicx}
\fi

\usepackage{amsmath}
\usepackage{amssymb}

\newtheorem{lemma}{Lemma}

\newtheorem{proposition}{Proposition}

\usepackage{algorithm, multirow, xcolor}
\usepackage{algorithmicx}
\usepackage{algpseudocode}

\usepackage{array}

\usepackage{url}

\usepackage{booktabs}
\usepackage{cancel}

\begin{document}
%
\title{Learning Efficient Flocking Control based on Gibbs Random Fields}
%
%
%

\author{Dengyu Zhang$^{*}$, Chenghao Yu$^{*}$, Feng Xue, and Qingrui Zhang
\thanks{This work is supported by supported by the Key-Area Research and Development Program of Guangdong Province under Grant 2024B1111060003, in part by the National Nature Science Foundation of China under Grant 62103451, Guang Dong Basic and Applied Basic Research Foundation  under Grant 2024A1515012408, and Shenzhen Science and Technology Program JCYJ20220530145209021. (Corresponding author: Qingrui Zhang, zhangqr9@mail.sysu.edu.cn)}
\thanks{$^{1}$All Authors are with School of Aeronautics and Astronautics, Shenzhen Campus of Sun Yat-sen University, Shenzhen 518107, P.R. China. $^{*}$Equal contributions.  }%

}
%
%

\markboth{IEEE Robotics and Automation Letters. 
Manuscript.
}
{Zhang \MakeLowercase{\textit{et al.}}: Learning Efficient Flocking Control based on Gibbs Random Fields} 

%



\maketitle

\begin{abstract}
Flocking control is essential for multi-robot systems in diverse applications, yet achieving efficient flocking in congested environments poses challenges regarding computation burdens, performance optimality, and motion safety. This paper addresses these challenges through a multi-agent reinforcement learning (MARL) framework built on Gibbs Random Fields (GRFs). With GRFs, a multi-robot system is represented by a set of random variables conforming to a joint probability distribution, thus offering a fresh perspective on flocking reward design. A decentralized training and execution mechanism, which enhances the scalability of MARL concerning robot quantity, is realized using a GRF-based credit assignment method. An action attention module is introduced to implicitly anticipate the motion intentions of neighboring robots, consequently mitigating potential non-stationarity issues in MARL. The proposed framework enables learning an efficient distributed control policy for multi-robot systems in challenging environments with {
success rate around $99\%$}, as demonstrated through thorough comparisons with state-of-the-art solutions in simulations and experiments. Ablation studies are also performed to validate the efficiency of different framework modules.

\end{abstract}



\IEEEpeerreviewmaketitle

\section{Introduction}
%
%
%
%
\IEEEPARstart{M}{ulti-robot} systems have shown great advantages over a single robot in terms of efficiency, robustness, and flexibility. In multi-robot systems, flocking is one of the fundamental behaviors to support their applications in many tasks, \emph{e.g.}, 
search-and-rescue operations\cite{tolstaya_learning_2020} and communication services\cite{bejaoui_qos-oriented_2020}. 
The burgeoning applications in diverse tasks put great demands on computational efficiency, performance optimality, and motion safety for multi-robot flocking.
In comparison with centralized flocking, a distributed solution is computationally more efficient\cite{park_online_2022}. However, it remains a challenge to simultaneously ensure efficiency, optimality, and safety for distributed large-scale multi-robot flocking in congested environments.

Tremendous algorithms have been proposed for distributed flocking control in past decades, many of which are bio-inspired methods stemming from the seminal work by Reynolds \cite{reynolds_flocks_1987}. Bio-inspired methods are rule-based, for instance, separation rules for collision avoidance, cohesion rules for robot congregation, and alignment rules for motion consensus \cite{olfati-saber_flocking_2006, roy_neural_2020}.
Typically, those rules in bio-inspired methods are designed based on artificial potential fields or certain velocity vectors \cite{vasarhelyi_optimized_2018, guo_collision-free_2023, mcguire_Viscoelastic_2022}. 
Bio-inspired methods are computationally efficient, but fall short of meeting various performance optimality requirements, \emph{e.g.} flocking order and {motion} tracking, \emph{etc}. Their collision avoidance behaviors will experience significant performance degeneration with the increase of robot quantity in congested environments, thus posing a safety issue.

In optimization-based methods, such as model predictive control (MPC), multi-robot flocking is formulated as an optimization problem, which is resolved online in an iterative way \cite{soria_predictive_2021}. Flocking control by MPC is, therefore, guaranteed to be optimal to some extent. Despite their advantages in performance guarantee, optimization-based methods are computationally expensive in general \cite{soria_distributed_2022}. To reduce the computational requirements, several distributed MPC (DMPC) methods have been proposed for flocking control \cite{soria_distributed_2022, lyu_multivehicle_2021
}. However, DMPC is perplexed by the decision mutual influence issue, in which the optimal decision of one robot depends on the optimization results of neighbor robots. 
The computation burden of DMPC is also proportional to the complexity of objective functions and system dynamics.
In addition, all MPC-based flocking control methods are model-based, so their performance in real systems is heavily affected by model accuracy.

\begin{figure}[tbp]
    \centering
    \includegraphics[width=1\linewidth]{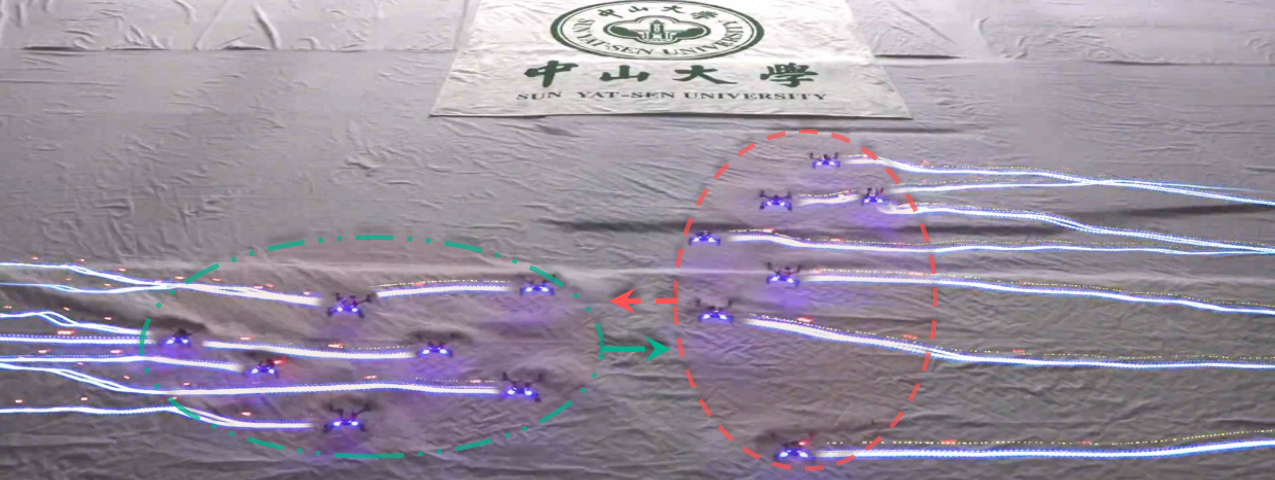}
    \caption{{Two flocks, each with 7 drones, move in opposite directions and avoid collisions in shared space using the proposed RL-based flocking controller.
    }}
    \label{fig:top_figure}
\end{figure}

Learning-based methods have recently emerged as a promising alternative to potentially achieve efficient, optimal, and safe flocking control in complex scenarios \cite{roy_neural_2020}.  In particular, reinforcement learning (RL) is capable of learning a policy by maximizing a return function using data samples collected through interactions with environments. The learning process is performed offline, so no more online optimization is needed. Hence, RL-based flocking control is potentially more time-efficient than optimization-based methods. Both the optimality and safety metrics are ensured by maximizing a specially designed reward function at the training stage. 
However, multi-robot flocking is much more challenging for RL than single-robot control. Firstly, robots in flocking need to actively interact with their surroundings, including neighboring robots and local environments, which significantly increases the difficulty of the reward design.  Secondly, the number of robots in flocking tends to change, so the current centralized training mechanism cannot be applied. Thirdly, the decision policies of robots, which keep updating at training, have a mutual influence on one another, leading to the nonstationarity issue.

In this paper, we aim to develop an RL-based algorithm for distributed multi-robot flocking control to ensure efficiency, optimality, and safety in congested environments. In the first contribution, we present a multi-agent reinforcement learning framework based on Gibbs random fields (GRFs) for flocking control. The maximization objective of the RL-based flocking is, therefore, modeled by a set of pairwise energy functions. The interaction between robots is modeled by undirected edges on a graph, while the robot-to-environment interaction is characterized using unary energy functions. An energy normalization technique is designed to reduce the nonstationarity issue caused by the dynamic changes in the interaction topology among robots at training.

In the second contribution, a decentralized training and decentralized execution (DTDE) mechanism is developed for distributed flocking control policy learning in light of a GRF-based credit assignment solution. 
The overall flocking objective is decomposed into a series of individual reward functions for each robot according to their contributions to return maximization, so the policy optimization is conducted in a distributed manner for every robot using their own rewards. The learning process is, therefore, scalable with the change of robot quantity in flocking. 

In the third contribution, we present an action attention module, which enables robots to implicitly anticipate the motion intention of neighbors. The implicit motion intention anticipation is achieved via exchanging previous action distribution with neighbors, in which a scaled dot-product attention mechanism is implemented \cite{vaswani_attention_2017}. With the action attention mechanism, a robot can make proactive reactions to the potential motion changes in its neighbor robots. The efficiency of the action attention is verified through simulations and experiments at various conditions.

%

\section{Related works}

Recently, GRFs have been applied to coordinating multi-robot behaviors \cite{yu_grf-based_2024,rezeck_chemistry-inspired_2022,zhu_heuristic_2024}. In GRFs, robots are represented by random variables that adhere to a joint distribution \cite{guo_collision-free_2023}. The interactions among robots are characterized by specific energy functions. The primary objective of GRF-based flocking control is to determine the optimal control actions by maximizing a posteriori distribution.  However,  GRF-based flocking control is computationally intensive due to the action optimization process. The computation burdens are partially alleviated by using a mean-field approximation technique for the case of small-scale flocking  \cite{koller_probabilistic_2009, fernando_online_2021}.  {Additionally, a heuristic predictive flocking control (HPFC) has been proposed to further decrease the computational demands in GRF-based flocking control by leveraging prior knowledge \cite{zhu_heuristic_2024}}. 
Despite these advancements, the existing GRF-based flocking control methods struggle to scale effectively with an increasing number of robots. Intensive communication remains essential during the online inference process, increasing the difficulty of real-time implementation.

As a promising alternative, multi-{agent} reinforcement learning (MARL) offers potentially more adaptable and scalable solutions regarding computational burdens, performance optimality, and motion safety. Early works primarily focused on direct applications of classical reinforcement learning algorithms, \emph{e.g.}, Q-learning 
\cite{hung_q-learning_2017}, and deep deterministic policy gradient \cite{wang_deep_2018}, \emph{etc}. These approaches only demonstrated their effectiveness in small-scale flocking scenarios with fewer than ten robots. To achieve better performance for large-scale flocking, a supervised learning framework is introduced \cite{roy_neural_2020}, which leverages centralized MPC as a teacher model.  Other modifications, such as 
graph attention mechanisms \cite{xiao_graph_2023}, have further enhanced the performance of MARL in terms of generalization and robustness. 
However, in these works, the coordination of robots depends on pre-defined training strategies without online intention recognition, reducing their flexibility in handling unpredictable environments.



\section{Preliminaries}\label{sec:prelim}

\subsection{Gibbs random field}\label{sec:gibbs-random-field}
A Gibbs random field (GRF) is represented by an undirected graph $\mathcal G = (\mathcal V, \mathcal E)$, where $\mathcal V = \{1,\ldots,n\}$ is the random variable set and $\mathcal{E} = \{(i,j)\in \mathcal{V} \times \mathcal{V} | j\in \mathcal{N}_i, i\in\mathcal{V}\}$ denotes the dependence among variables with $\mathcal{N}_i$ as the neighbor set of a variable $i$ \cite{koller_probabilistic_2009}. In a GRF, the joint probability of random variables $X = \{X_v\}_{v\in\mathcal V}$ is delineated as 
\begin{equation}
P(X) = \left.\prod\nolimits_{c \in \mathcal{C}\text{, }X_c\subseteq X} \phi_c (X_c)\right/Z
	\label{gibbs_def}
\end{equation}
where $\mathcal C \subseteq \mathcal V$ is a clique set, $\phi_c$ is a potential function for a clique $c\in \mathcal C $, $Z = \sum_{X} \prod_{c \in \mathcal{C}\text{, }X_c\subseteq X} \phi_c (X_c)$ is a normalization constant. All robots in a clique $c$ are neighbors of one another.

When $P(X)$ obeys a Gibbs distribution, $\phi_c(X_c)$ has an exponential form of  $\phi_c(X_c) = \exp \{ - \psi_c(X_c) \}$ with $\psi_c(X_c)$ interpreted as free energy, so \eqref{gibbs_def} is thus written as
\begin{equation}
	P(X) 
  = \left.\exp \{ -H(X) \}\right/Z
	\label{gibbs_distribution}
\end{equation}
where $H(X) = \sum_{c\in \mathcal{C}} \psi_c(X_c)$. The derivation of Gibbs formulae defining a GRF is from the physical fact that the interaction between particles is described by a potential. In flocking, each robot can be treated as a random particle, so all robots as a group satisfy a certain Gibbs distribution.

\subsection{Partially observable Markov decision process}
In distributed flocking control, global information is unobservable for each robot. Hence, the coordination of robots in flocking is modeled as a decentralized partially observable Markov decision process (Dec-POMDP) \cite{oliehoek_concise_2016}, which is 
defined by a tuple $\langle \mathcal V, \mathcal S, \{\mathcal A_i\}, \mathcal T, \{\mathcal O_i\}, \mathcal Z, r, \gamma \rangle$, 
where $\mathcal V = \{1, \dots, n\}$ is the set of robots, $\mathcal S$ is the global state space, 
$\mathcal A_i$ is the action space of robot $i$, and $\mathcal O_i$ is the observation space for robot $i$, $\mathcal T: \mathcal S \times \prod_{\forall i\in\mathcal V} \mathcal A_i \times \mathcal S \rightarrow [0,1]$ defines the transition probability from the current state to the next state, $\mathcal Z: \mathcal S \rightarrow \prod_{\forall i\in\mathcal V} \mathcal O_i$ is the observation function, $r: \mathcal S \times \prod_{\forall i\in\mathcal V} \mathcal A_i \rightarrow \mathbb{R}$ is a reward function, and $\gamma \in [0,1)$ is a discount factor. A joint policy for all robots in flocking is defined as $\boldsymbol{\pi}=\prod_{\forall i\in\mathcal V}\boldsymbol{\pi}_i$, where $\boldsymbol{\pi}_i\left(\left.\boldsymbol{a}_i\right| \boldsymbol{o}_i\right)$ is a policy for robot $i$ with $\boldsymbol{a}_i\in \mathcal A_i$ and $\boldsymbol{o}_i\in \mathcal O_i$.  A return function is defined as an accumulation of future rewards, namely $G_t = \sum_{l=0}^{\infty} \gamma^l r_{t+l}$,
where $r_t$ is the reward at the time step $t$. 
The objective of Dec-POMDP is to find a policy $\boldsymbol{\pi}^*$ that maximizes a value function  $V^{\boldsymbol{\pi}}(\boldsymbol{s}_t) = \mathbb E[G_t | \boldsymbol{s}_t]$ or an action-value function
$Q^{\boldsymbol{\pi}}(\boldsymbol{s}_t, \{\boldsymbol{a}_{i,t}\}_{i}^{n}) = \mathbb E[G_t | \boldsymbol{s}_t,\{\boldsymbol{a}_{i,t}\}_{i}^{n}]$.

\subsection{Robot dynamics and observations}
The robots in flocking are assumed to have homogeneous second-order dynamics given by $\dot{\boldsymbol p}_i = \boldsymbol v_i$, $\dot{\boldsymbol v}_i = \boldsymbol a_i$, 
where $\boldsymbol p_i$ is the position vector, $\boldsymbol v_i$ is the velocity vector, and $\boldsymbol{a}_i$ is the acceleration of robot $i$. For simplicity, the flocking behavior in two-dimensional space is considered, so $\boldsymbol p_i$, $\boldsymbol v_i$, $\boldsymbol{a}_{i} \in \mathbb{R}^{2}$. However, the proposed algorithm can be extended to a three-dimensional case with nearly no modifications.

In our proposed design{, robots can only change their accelerations, so the action space is represented by a set of acceleration vectors.} The action space is hybrid with $\boldsymbol a_i$ from either a discrete action set or a continuous policy set. Let $\mathcal{A}_i$ be the hybrid action space for robot $i$ with $ \mathcal{A}_i = \mathcal{A}_{i,d} \cup \mathcal{A}_{i,c}$, where $\mathcal{A}_{i,d}$ denotes the discrete action set and $\mathcal{A}_{i,c}$ is the continuous policy. The discrete action set $\mathcal{A}_{i,d}$ is obtained by 
\begin{equation*}
\mathcal{A}_{i,d} = \{\boldsymbol{a}_{i} = M \boldsymbol{e} | M \in \left\{M_1, \ldots, M_m\right\}, \boldsymbol{e} \in\left\{\boldsymbol{e}_1, \ldots, \boldsymbol{e}_{l}\right\} \}
\end{equation*}
where $M$ is the action magnitude, and $\boldsymbol{e}$ is the action direction.
The continuous policy set $\mathcal{A}_{i,c}$ contains two simple rules to enhance motion smoothness, so $\mathcal{A}_{i,c}=\{\boldsymbol{a}_{i,s}, \boldsymbol{a}_{i,f}\}$. The first rule $\boldsymbol{a}_{i, s} = -\boldsymbol{v}_i$ is used to decelerate a robot in a more smooth manner. The second rule $\boldsymbol{a}_{i, f}=\boldsymbol{v}_c - \boldsymbol{v}_i$ regulates the velocity tracking performance, where $\boldsymbol{v}_c$ is the desired flocking velocity. Other rules could be added to $\mathcal{A}_{i,c}$, if necessary.
\begin{figure}[htbp]
\centering
\includegraphics[width=0.75\linewidth]{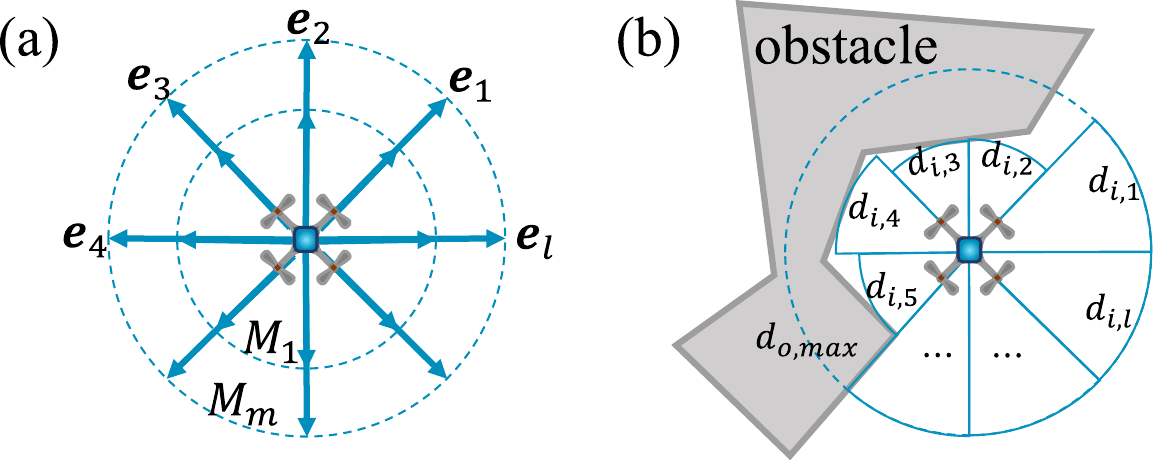}
    \caption{{(a): Discrete action set $\mathcal{A}_{i,d}$ indicating acceleration vectors that robots can choose. (b): Obstacle observation $\boldsymbol{o}_{i,o}$ consists of the distances to obstacles in $l$ evenly divided sectors. The distances are defined as the minimum radius of the sector that doesn't cover any obstacles.}
    }
	\label{fig:perception}
\end{figure}
The neighborhood for robot $i$ is defined by $\mathcal{N}_i = \{j | \Vert \boldsymbol{p}_i - \boldsymbol{p}_j \Vert \leq d_{in}, j\in \mathcal{V}\; \& \; j\neq i\}$, where $d_{in}$ is the interaction range. 
{The observation space of a robot is denoted as $\boldsymbol{o}_i = \{\boldsymbol{o}_{ii}, \{\boldsymbol{o}_{ij}\}_{j\in \mathcal{N}_i}, \{\boldsymbol{A}_{j}\}_{j\in \mathcal{N}_i}, \boldsymbol{o}_{io}\}$, where $\boldsymbol{o}_{ii}= \{\boldsymbol p_i, \boldsymbol v_i, \boldsymbol{p}_c, \boldsymbol{v}_c\}$  contains position and velocity of robot $i$ ($\boldsymbol p_i, \boldsymbol v_i$), and the desired position and velocity of the flock ($\boldsymbol{p}_c, \boldsymbol{v}_c$)}. $\boldsymbol{o}_{ij} = \{\boldsymbol p_j - \boldsymbol p_i, \boldsymbol{v}_j - \boldsymbol{v}_i\}$ denotes the relative state of a neighbouring robot $j$ to robot $i$. $\boldsymbol{A}_{j}$ is the action distribution of a neighbouring robot $j$ at the last timestep, and $\boldsymbol{o}_{io}= \{d_{i,1}, ..., d_{i,l}\}$ specifies the distances to obstacles in $l$ different directions. The obstacles are detected in a similar way as LiDAR. It is assumed that robot $i$ has the omnidirectional visibility of surrounding obstacles, so the obstacle detection has a span of $2\pi$ (360 degrees). We divide the detection span evenly into $l$ sectors, so $d_{i, k}$ represents the closest obstacle in the $k$-th sector with $d_{i, k}<d_{o, max}$, where $d_{o, max}$ specifies the maximum detection range. The obstacle detection mechanism is shown in Fig. \ref{fig:perception}(b).
\section{Methodology}\label{sec:method}

{In this section, an RL-based algorithm is designed for flocking control in congested environments. The optimal control is learned by maximizing a return function of Dec-POMDP, which corresponds to minimizing the total free energy in the GRFs. 
Hence, different free energy terms are presented, which is followed by the construction of a GRF.
A credit assignment mechanism is presented for decentralized training. An action attention structure is eventually developed for the policy network to achieve implicit motion intention anticipation. }

\subsection{Flocking reward}
\label{sec:flocking_reward}
The interactions in flocking are modeled as energy functions in a GRF. Two types of energy functions are rendered: pairwise energy for robot-to-robot interactions and unary energy for robot-to-environment interactions, so the total free energy is 
\begin{equation}
\label{eq:singleton_pairwise_potential}
    H(X) = \sum\nolimits_{i \in \mathcal V} \psi_i(\boldsymbol{o}_{i}, \boldsymbol{a}_{i}) + \sum\nolimits_{(i,j)\in \mathcal E} \psi_{ij}(\boldsymbol{o}_{i}, \boldsymbol{o}_{j})
\end{equation}
where $\psi_i(\boldsymbol{o}_{i}, \boldsymbol{a}_{i})$ is unary energy, $\psi_{ij}(\boldsymbol{o}_{i}\boldsymbol{o}_{j})$ is pairwise energy. In \eqref{eq:singleton_pairwise_potential}, $\boldsymbol{o}_i$ are regarded as random variables in GRFs, while $\boldsymbol{a}_i$ are taken as parameters.


Pairwise energy $\psi_{ij}$ contains position alignment energy and velocity alignment energy to ensure cohesion, separation, and alignment in flocking, which is defined as 
\begin{equation}
	\psi_{ij}(\boldsymbol{o}_{i}, \boldsymbol{o}_{j}) = \psi_{ij,p}(\boldsymbol{o}_{i}, \boldsymbol{o}_{j}) + \psi_{ij,v}(\boldsymbol{o}_{i}, \boldsymbol{o}_{j})
\end{equation}
where $\psi_{ij,p}$ is the position alignment energy, $\psi_{ij, v}$ is the velocity alignment energy. Here we choose the Morse potential \cite{morse_diatomic_1929} as the position alignment energy to ensure flocking cohesion and separation, so 
\begin{equation} 
\psi_{ij,p} (\boldsymbol{o}_{i},\boldsymbol{o}_{j}) = c_{p1}\left[1 - \exp(-c_{p2}(d_{ij}-d_{r}))\right]^2 - c_{p1}
\label{position_alignment_energy}
\end{equation}
where $c_{p1}, c_{p2} > 0$ are coefficients, $d_{ij}=\Vert \boldsymbol p_i - \boldsymbol p_j \Vert$ is the distance between two robots with $d_{r}$ as the desired value. The energy $\psi_{ij,p}$  has three properties: 1) $\psi_{ij,p}$ is minimum at $d_{ij}=d_{r}$; 2) $\psi_{ij,p}(d_{r}) = -c_{p1} < 0$; 3) $\forall \epsilon \in (0$, $ d_{r})$, $\left.\psi_{ij,p}\right|_{d_{ij}=d_{r}-\epsilon} > \left.\psi_{ij,p}\right|_{d_{ij}=d_{r}+\epsilon}$.

The velocity alignment energy $\psi_{ij,v}$ is designed as
\begin{equation} 
\psi_{ij,v} (\boldsymbol{o}_{i}, \boldsymbol{o}_{j}) = c_v \max\left(0, -\boldsymbol v_{ji}^\top \boldsymbol p_{ji} / \Vert \boldsymbol p_{ji} \Vert^2 \right)
\label{velocity_alignment_energy}
\end{equation}
where $c_v>0$ is a coefficient, $\boldsymbol v_{ji}=\boldsymbol{v}_j - \boldsymbol{v}_i$ is the relative velocity, and $\boldsymbol p_{ji}=\boldsymbol{p}_j - \boldsymbol{p}_i$ is the relative position between robots. The velocity alignment energy is designed to minimize the velocity in the direction of relative position. 

Unary energy $\psi_i$ is given by a sum of several terms as
\begin{equation}
	\psi_i(\boldsymbol{o}_{i}, \boldsymbol{a}_i) = \psi_{i,k} + \psi_{i,c} + \psi_{i,t} + \psi_{i,o} + \psi_{i,b}
\end{equation}
where $\psi_{i,k}$ is the energy for motion smoothness, $\psi_{i,c}$ is the control optimization energy, $\psi_{i,t}$ is the position tracking energy, $\psi_{i,o}$ is the obstacle avoidance energy, and $\psi_{i,b}$ is the collision aversion energy. 

The motion smoothness energy $\psi_{i,k}$ is beneficial to reducing oscillations, which is formulated as $\psi_{i,k}(\boldsymbol{o}_{i}) = c_k {\Vert \boldsymbol{v}_i - \boldsymbol{v}_c\Vert^2}$, 
where $c_k>0$ is a coefficient, $\boldsymbol{v}_c$ is the desired velocity{ of the flock}. 
The control optimization energy $\psi_{i,c}(\boldsymbol{o}_i)$ is defined as $\psi_{i,c}(\boldsymbol{a}_i) = c_{c}\boldsymbol{a}_i^2$.
The position tracking energy $\psi_{i,t}$ encourages a robot  to approach the flocking {desired} position, which is given by $\psi_{i, a}(\boldsymbol{o}_{i}) = c_{t1} (\Vert \boldsymbol p_i - \boldsymbol p_c \Vert - c_{t2})$, 
where $c_{t1}$, $c_{t2}$ are constants, and $\boldsymbol p_c$ is the {desired position of flock}. 
The obstacle avoidance energy $\psi_{i,o}$ is designed similarly to the Morse potential, but only the repulsive part is left.
\begin{equation}
    \psi_{i,o}(\boldsymbol{o}_i) = c_{o1} \left\{ 1 - \exp [-c_{o2} \min (0, d_{oi} - d_{or})] \right\}^2
\label{eq:obstacle_energy}
\end{equation}
where $c_{o1}, c_{o2} > 0$ are coefficients, $d_{oi} = \Vert \boldsymbol p_{o,min} - \boldsymbol p_{i} \Vert$ is the distance to nearest obstacle, $\boldsymbol p_{o,min}$ is the position of the nearest obstacle, $d_{or}$ is a constant indicating the reaction distance to obstacle. 
A collision aversion energy $\psi_{i,b} (\boldsymbol{o}_i) $ is introduced to limit the velocity towards obstacles. 
\begin{equation}
\psi_{i,b} (\boldsymbol{o}_i) = c_{b} \max\left(0, -\boldsymbol v_{oi}^\top \boldsymbol p_{oi} / \Vert \boldsymbol p_{oi} \Vert^2 \right)
\label{brake_energy}
\end{equation}
where $c_{b}$ is a coefficient, $\boldsymbol p_{oi} = \boldsymbol p_{o,min} - \boldsymbol p_{i}$ is the relative position of the nearest obstacle, $\boldsymbol v_{oi} = \boldsymbol v_{o,min} - \boldsymbol v_{i}$ is the relative velocity of the nearest obstacle.


\begin{figure}[htbp]
    \centering
    \includegraphics[width=0.6\linewidth]{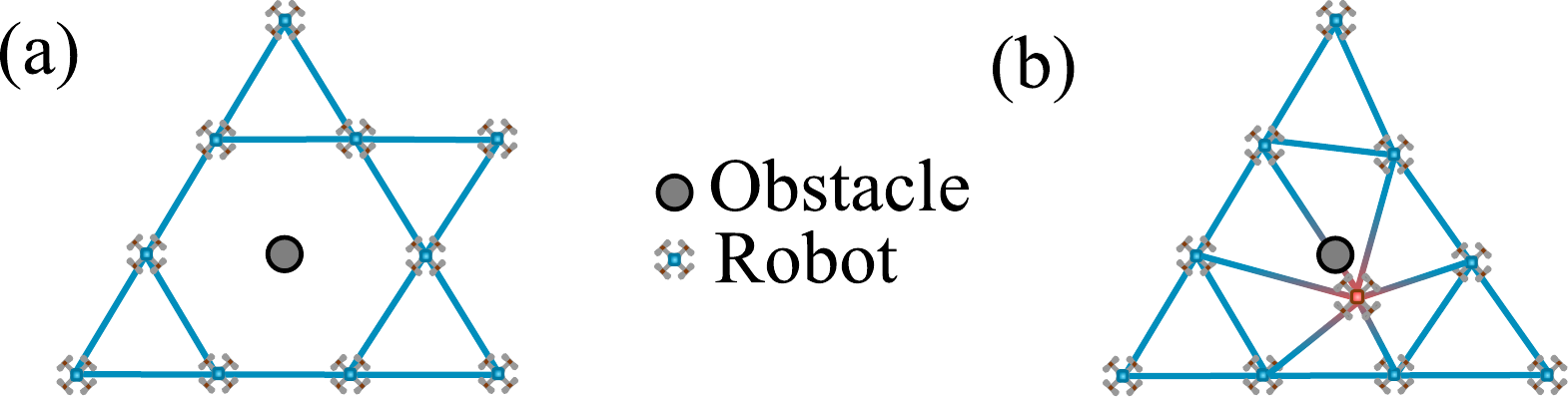}
    \caption{{Configurations with (a) 14 edges, (b) 18 edges. The red robot in (b) tends to collide with obstacles due to attractions from extra neighbors.}}
    \label{fig:energy_normalization}
\end{figure}
{The interaction edges in $\mathcal{E}$ might change due to the variation caused by distance change of the neighbor robots of robot $i$, so the number of items to be summed up in \eqref{eq:singleton_pairwise_potential} can dynamically change. Although more edges usually lead to better cohesion, more edges might degenerate safety and collision avoidance performance, as shown in Fig. \ref{fig:energy_normalization}}. To balance cohesion and safety, the total free energy $H(X)$ in (\ref{eq:singleton_pairwise_potential}) is modified to be 
\begin{equation}
\color{red}
\label{eq:normalized_energy}
    H(X) = H_u(X) + H_p(X)
\end{equation}
{where $H_u(X) = \sum_{i \in \mathcal V} \psi_i(\boldsymbol{o}_{i}, \boldsymbol{a}_i)$ is the unary energy, $H_p(X) = \frac{\vert \mathcal{V} \vert}{\vert \mathcal{E} \vert}\sum_{(i,j)\in \mathcal{E}} \psi_{ij}(\boldsymbol{o}_{i}, \boldsymbol{o}_{j})$ is the normalized pairwise energy,} $\vert \cdot \vert$ denotes the cardinal number of a set.
The reward function for robot flocking is, therefore, designed to be 
\begin{equation}
\label{eq:collective_reward}
	r = \exp[-{H(X)}]  \propto P(X)
\end{equation}

\subsection{Credit assignment for decentralized training}
\label{sec:gibbs_payoff_distribution}

The flocking reward presented in Section IV-A is formulated from a centralized perspective, relying on global information. However, in a distributed system, robots only have access to information from their immediate neighbors, making it impossible for each robot to infer the global flocking reward solely from its local observations. 
Most RL algorithms assume a stationary environment, but the assumption is violated when the observable information cannot provide sufficient information for optimal decision-making \cite{wong_deep_2023}. 
To mitigate this issue, we propose a credit assignment method that constructs a series of local rewards based on each robot's local observations, leading to a DTDE mechanism. 

In the global reward function \eqref{eq:collective_reward}, the couplings among robots arise from the pairwise energy terms in the global energy \eqref{eq:normalized_energy}. Hence, we introduce a decoupled pairwise energy.

\begin{equation}
\label{eq:decoupedPairwise}
    \hat{H}_p = \sum\nolimits_{i \in \mathcal{V}} \frac{1}{|\mathcal{N}_i|} \sum\nolimits_{j \in \mathcal{N}_i} \psi_{ij} (\boldsymbol{o}_{i}, \boldsymbol{o}_{j})
\end{equation}
where $\hat{H}_p$ is a decoupled pairwise energy which approximates the pairwise energy $H_p$ in \eqref{eq:normalized_energy}.

{The approximation is valid because $H_p$ and $\hat{H}_p$ share the same minimum condition and minimum value, which is proved in Proposition \ref{Thm:same_minima}.}

\begin{proposition}
\label{Thm:same_minima}
The decoupled pairwise energy $\hat{H}_p $ given in \eqref{eq:decoupedPairwise} shares the same minimum condition and minimum value with the pairwise energy given in \eqref{eq:normalized_energy}.
\end{proposition}

\begin{IEEEproof}
{According to the first property of Eq. \eqref{position_alignment_energy}, the position alignment energy $\psi_{ij,p}$ reaches a minimum value of $-c_{p1}$ when  at $\Vert \boldsymbol{p}_{ji} \Vert = d_{r}$. The velocity alignment energy $\psi_{ij,v} (\boldsymbol{o}_{i}, \boldsymbol{o}_{j})$ reaches a minimum value of $0$ when $\boldsymbol v_{ji}^\top \boldsymbol p_{ji}\geq 0$. Hence, the minimum point for $H_p$ and $\hat{H}_p$ must satisfy the conditions that 1) $\Vert \boldsymbol{p}_{ji} \Vert = d_{r}$ and 2) $\boldsymbol v_{ji}^\top \boldsymbol p_{ji}\geq 0$, $\forall (i,j) \in \mathcal{E}$. The second condition holds if $\boldsymbol v_{ji}^\top \boldsymbol p_{ji}> 0$, $\boldsymbol v_{ji} \perp \boldsymbol{p}_{ji}$, or $\boldsymbol v_{ji}=\boldsymbol{0}$, $\forall (i,j) \in \mathcal{E}$. When $\boldsymbol v_{ji}^\top \boldsymbol p_{ji}> 0$, it implies that the angle between the velocity vector $\boldsymbol{v}_{ji}$ and the position vector $\boldsymbol{p}_{ji}$ is smaller than $\frac{pi}{2}$, namely $\angle\left(\boldsymbol{v}_{ji}, \boldsymbol {p}_{ji}\right)<\frac{\pi}{2}$. In this case, two robots would move towards each other, leading to an increase in $\psi_{ij,p}$. Hence, the case of $\boldsymbol v_{ji}^\top \boldsymbol p_{ji}> 0$ is not a minimum condition for $H_p$ or $\hat{H}_p$. When $\boldsymbol v_{ji} \perp \boldsymbol{p}_{ji} =0$, one robot would circle around another robot, by which the in-between distance could also be kept. Hence, both $H_p$ and $\hat{H}_p$ reaches a minimum value only if $\Vert \boldsymbol{p}_{ji} \Vert = d_{r}$ and $\boldsymbol v_{ji}^\top \boldsymbol p_{ji} = 0$. 
The minimum value of them is $\sum_{i \in \mathcal{V}} -c_{p1} = -|\mathcal{V}|c_{p1}$
}
\end{IEEEproof}

{The minimum condition in Proposition \ref{Thm:same_minima} corresponds to configurations that every robot keeps the desired distance to its neighbors. A special case is the $\alpha$-lattice in \cite{olfati-saber_flocking_2006}, which is realizable when $d_{in}$ is properly selected, \emph{e.g.}, $d_r \leq d_{in} < \sqrt{2}d_{r}$ in a 2D space.}
Note that the circling motion due to $\boldsymbol v_{ji} \perp \boldsymbol{p}_{ji}$ is highly unlikely to happen for large-scale flocking. Other than the pairwise energy, there is also unary energy, such as motion smoothness $\psi_{i,k}(\boldsymbol{o}_{i})$ and position tracking {$\psi_{i, t}(\boldsymbol{o}_{i})$}, which would inhibit the occurrence of circling motion.

{The total pairwise $H_p$ and the decoupled pairwise energy $\hat{H}_p$ share the same minimum conditions, so we can replace the total pairwise energy $H_p$ with $\hat{H}_p$ in \eqref{eq:normalized_energy}, leading to a series of local rewards $\{r_i\}$ for each robot.
}
\begin{equation}
\label{eq:local_reward}
    r_i = \exp{\left[-\psi_i(\boldsymbol{o}_{i}, \boldsymbol{a}_i) - \frac{1}{|\mathcal{N}_i|}\sum\nolimits_{j \in \mathcal{N}_i} \psi_{ij}(\boldsymbol{o}_{i}, \boldsymbol{o}_{j})\right]}
\end{equation}

\subsection{Action attention structure}
\label{sec:action_attention}

In this subsection, an action attention structure is presented to implicitly anticipate the motion intention of neighboring robots. The action attention structure is inspired by the mean field theory, in which a joint coupled distribution could be locally approximated by a factored distribution in terms of Kullback–Leibler (KL) divergence. 
\begin{lemma}[Mean-field approximation \cite{koller_probabilistic_2009}]
\label{lemma:mean_field} Suppose $P(X)$ is a joint distribution and $Q(X)$ is a class of distributions specified as a product of independent marginals, namely $Q(X)=\prod_{i\in \mathcal{V}} Q(X_i)$. The factored distribution $Q(X)$ is a local optimal approximation of  $P(X)$ in terms of $\mathcal{D}_{KL}\left[Q(X)\Vert P(X)\right]$, if and only if
\begin{equation}
Q(X_i) = \frac{1}{Z_i}\exp\left\{ \mathbb{E}_{X_{-i}\sim Q}[\ln P(X_i | X_{-i})] \right\}
\end{equation}
where $Z_i$ is a normalization constant, $X_{-i}$ denotes all the random variables in $X$ except $X_i$, and $\mathcal{D}_{KL}\left[Q(X)\Vert P(X)\right]$ denotes the KL divergence between two distributions with $\mathcal{D}_{KL}\left[Q(X)\Vert P(X)\right]=\sum_{X}Q(X)\ln\left[\frac{Q(x)}{P(x)}\right]$. 
\end{lemma}
\begin{IEEEproof}
The proof of lemma \ref{lemma:mean_field} can be found in \cite{koller_probabilistic_2009}. 
\end{IEEEproof}

The modified energy in \eqref{eq:normalized_energy} leads to a joint distribution $P(X) \propto \exp[-H(X)]$. 
Based on Lemma \ref{lemma:mean_field}, the optimal choice $Q(\boldsymbol{o}_i)$ for robot $i$ to approximate $P(X)$ has the following formula.
\begin{equation}
    Q(\boldsymbol{o}_i) \propto \exp \left[-\psi_i - \frac{|\mathcal V|}{|\mathcal E|}\sum_{j \in \mathcal N_i} \sum_{\boldsymbol{o}_j \in \mathcal{O}_j} Q(\boldsymbol{o}_j) \psi_{ij}\right]
 \label{eq:meanfield}
\end{equation}
In \eqref{eq:meanfield}, a weighted sum of neighbor action distribution $Q(\boldsymbol{o}_j)$ is required to obtain $Q(\boldsymbol{o}_j)$.

To obtain such weights $\psi_{ij}$ in the policy network, we introduce the scaled dot-product attention module. In scaled dot-product attention, the network inputs contain queries $\boldsymbol{q}_j\in\mathbb{R}^{d_{k}\times 1}$, keys $\boldsymbol{k}_j\in\mathbb{R}^{d_{k}\times 1}$, and values $\boldsymbol{l}_j\in\mathbb{R}^{d_{l}\times 1}$ \cite{vaswani_attention_2017}. The attention weights are computed by a softmax of the dot product of queries and keys.
$\mathrm{Attention}(\boldsymbol{Q}, \boldsymbol{K}, \boldsymbol{L}) = \mathrm{softmax}\Big(\boldsymbol{Q}^\top \boldsymbol{K}/\sqrt{d_k}\Big)\boldsymbol{L}$,
where $\boldsymbol{Q}=\left[\boldsymbol{q}_1,\ldots,\boldsymbol{q}_m\right]$, $\boldsymbol{K}=\left[\boldsymbol{k}_1,\ldots,\boldsymbol{k}_m\right]$, $\boldsymbol{L}=\left[\boldsymbol{l}_1,\ldots,\boldsymbol{l}_l\right]$.

The neighborhood information $\boldsymbol{o}_{ij}$ is firstly embedded into $\boldsymbol{e}_j$ by an embedding linear layer with a rectified linear unit (ReLU) activation function. The query $\boldsymbol{q}_j$ and value $\boldsymbol{l}_j$ are then extracted by two different linear layers from $\boldsymbol{e}_j$. The action distribution $\boldsymbol{A}_{j}$ of the neighbor robot $j$ is taken as the key $\boldsymbol{k}_j$, so $\boldsymbol{k}_j=\boldsymbol{A}_{j}$. The attention weight $\alpha_j$ is calculated by using a dot product of $\boldsymbol{q}_j$ and $\boldsymbol{A}_{j}$, namely $\alpha_j = \boldsymbol{q}_j^\top \boldsymbol{A}_{j} / \sqrt{|\boldsymbol{q}_j|}$.

\begin{figure}[htbp]
\centering
\includegraphics[width=0.9\linewidth]{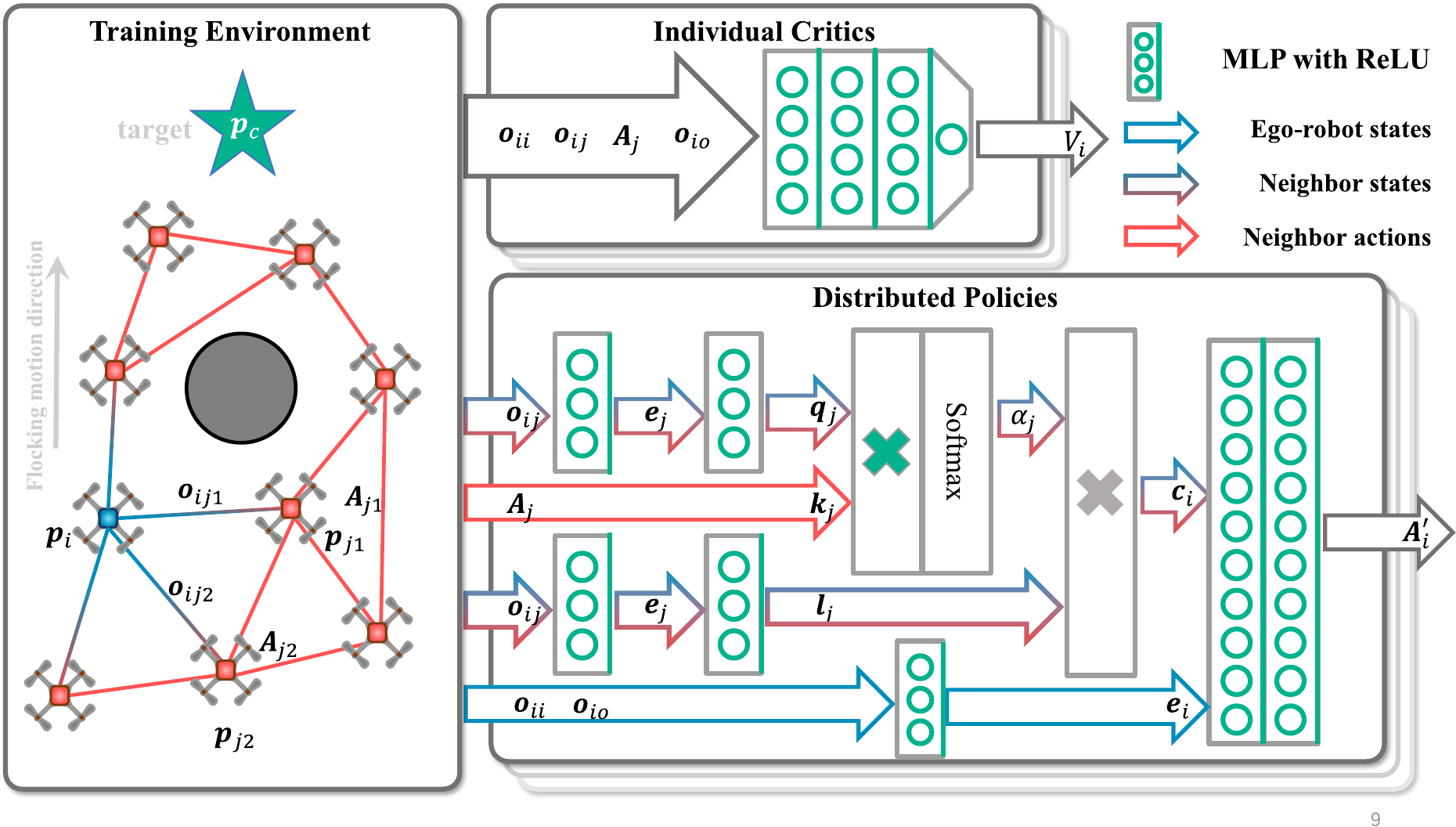}
\caption{{An action attention structure is designed for distributed policies. The attention weight $\alpha_j$ evaluating the importance of each neighbor is computed by both neighbor observation $\boldsymbol{o}_{ij}$ and neighbor action distribution $\boldsymbol{A}_j$. 
}}
\label{fig:framework}
\end{figure}

In this structure, $\boldsymbol{q}_i$ can be interpreted as the awareness of each action. Thus, if an action input of a neighbor robot $j$ has a high probability (or larger action-state value), the corresponding attention weight will be high as well, implying robot $i$ should pay more attention to the high possible moves of its neighboring robots. Implicit intention approximation is therefore achieved based on the previous action distribution of neighbors.
The embedded feature $\boldsymbol{c}_i$ of neighbors is an attention-weighted sum of embedded value $\boldsymbol{l}_j$, so $\boldsymbol{c}_i = \sum\nolimits_{j \in \mathcal N_i} \alpha_j \boldsymbol{l}_j$.
Information of robot $i$ itself, $\boldsymbol{o}_{ii}$ and $\boldsymbol{o}_{io}$, is concatenated and embedded into $\boldsymbol{e}_i$ by a linear layer with ReLU activation function. Then, $\boldsymbol{e}_i$ and $\boldsymbol{c}_i$ are concatenated and fed into a multi-layer perceptron (MLP) with ReLU activation functions to generate the new action distribution $\boldsymbol{A}_{i}' = \mathrm{softmax}[MLP(\boldsymbol{e}_i, \boldsymbol{c}_i)]$ for robot $i$.
The action attention structure is shown in Fig. \ref{fig:framework}. 

\subsection{Decentralized policy optimization}
We choose the Proximal Policy Optimization (PPO) algorithm to solve the multi-robot policy learning problem, although {m}ost standard RL algorithms could be applied. Based on \eqref{eq:local_reward}, the state value function for each robot $i$ is approximated using a critic network $V_i^{\boldsymbol{\pi}_{i}}$, while the control policy $\boldsymbol{\pi}_{i}$ is constructed based on the action attention structure. 

The parameters of $V_i^{\boldsymbol{\pi}_{i}}$ are trained by minimizing a mean square error $L_{c,i} = \mathbb{E} [G_{t,i} - V_i^{\boldsymbol{\pi}_i}]^2$, where $V^{\boldsymbol{\pi}_i}$ is the state value estimated by the critic network. The policy network $\boldsymbol{\pi}_{i}$ is optimized by maximizing a surrogate objective $L_{a,i} = \mathbb{E} \{L_{cl,i} + c_e S[\boldsymbol{\pi}_i]\}$, 
where 
$c_e$ is entropy coefficient, $S[\boldsymbol{\pi}_i]$ is an entropy bonus, and $L_{cl, i}$ is a clipped surrogate objective based on general advantage estimation (GAE) \cite{schulman_proximal_2017}. 

At training, the critic networks and policy networks for all robots are optimized with a common learning rate $\beta$. Networks are updated per batch for $n_p$ epochs with minibatch size $n_m$. In every batch, data from $n_b$ episodes is collected. 

\section{Results}


\subsection{Training setup}
At the beginning of every episode, a random number of static obstacles (up to $n_{o, max}=50$) are randomly generated in a $15\mathrm{m} \times 15\mathrm{m}$ square arena. Robots are initially randomly generated in the same square arena with a random velocity (up to $v_{c, max}=0.4$) as shown in Fig. \ref{fig:cluster_crossing}. The desired position is initialized at the arena center with a random velocity (up to $v_{c, max}$). 
At every training episode, a robot would instantly stop for $t_c = 5$ seconds, if it collides with an obstacle. The discrete action set $\mathcal{A}_{i,d}$ in Fig. \ref{fig:perception} are chosen such as $M \in \{0.05, 0.1, 0.2, 0.5\}$ with $\boldsymbol{e}\in\left\{\boldsymbol{e}_1,\ldots,\boldsymbol{e}_l,\ldots\boldsymbol{e}_{32}\right\}$ with $\boldsymbol{e}_l$ has an azimuth angle of $\frac{l\pi}{16}$.
The policy network is trained for $50000$ episodes ($25$ million steps) and the total number of steps in every episode is $500$. Training parameters are $\gamma=0.99$, $\beta=0.0005$, $c_e=0.001$, $c_c=0.00001$, $c_{v} = 0.001$, $c_{k} = 0.02$, $c_{t1}=0.005$, $c_{t2}=3$, $c_{p1} = 0.03$, $c_{p2} = 1.5$, $c_{o1} = 0.03$, $c_{o2} = 2$, $c_{o3} = 0.01$, $n_b = 5$, $n_m = 256$.


\subsection{Framework validation}
\begin{figure}[htbp]
    \centering
    \includegraphics[width=1\linewidth]{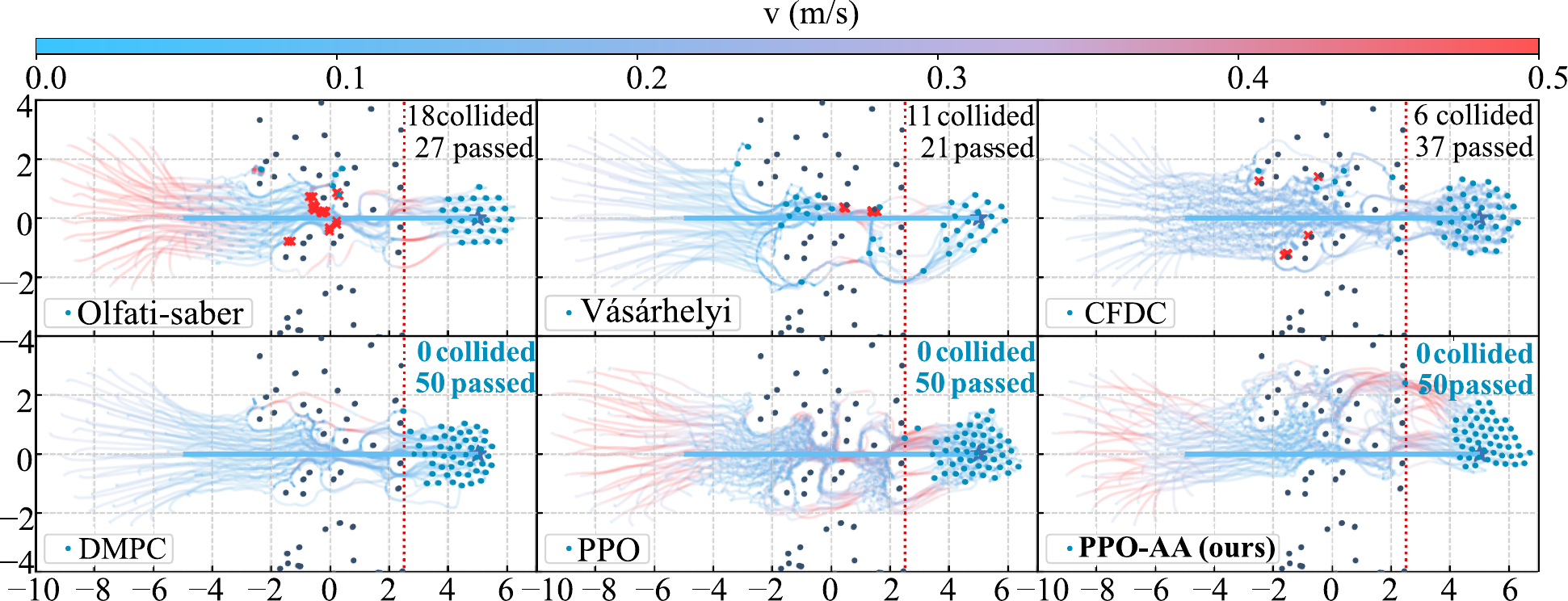}
    \caption{{A validation environment with $50$ obstacles, in which robots move from the left to the right by six different algorithms:} (a) Olfati-saber, (b) V\'as\'arhelyi, (c) CFDC, (d) DMPC, (e) PPO, (f) PPO-AA.}
    \label{fig:cluster_crossing}
\end{figure}

The proposed framework is evaluated in environments with various numbers of robots, \emph{e.g.} $10$, $30$, and $50$, and also different quantities of obstacles, \emph{e.g.}  $10$, $20$, and $50$. This setup is not used at training, so it is employed to test the efficiency and generalization of the proposed algorithm. 
A simulation case is shown in Fig. \ref{fig:cluster_crossing}. The proposed algorithm, which is termed as PPO-AA (PPO with action attention), is compared with state-of-the-art benchmarks, including PPO without action attention (PPO), {three rule-based methods (Olfati-saber's algorithm\cite{olfati-saber_flocking_2006}, Vásárhelyi's algorithm\cite{vasarhelyi_optimized_2018}, CFDC\cite{guo_collision-free_2023}), a distributed MPC method (DMPC) \cite{soria_distributed_2022}, and a GRF-based method (HPFC) \cite{zhu_heuristic_2024}.} The comparison is performed in terms of computation efficiency, flocking optimality, and motion safety.

All algorithms are repeatedly run $10$ times at each scenario with the same amount of robots and obstacles. For each run, the locations of all obstacles will be randomly generated. 
It implies that the algorithms are sufficiently evaluated at $90$ different setups in total. The mean and variance of each performance metric are summarized for fair evaluation.


\subsubsection{Computation efficiency}
\begin{figure}[htbp]
    \centering
    \includegraphics[width=1\linewidth]{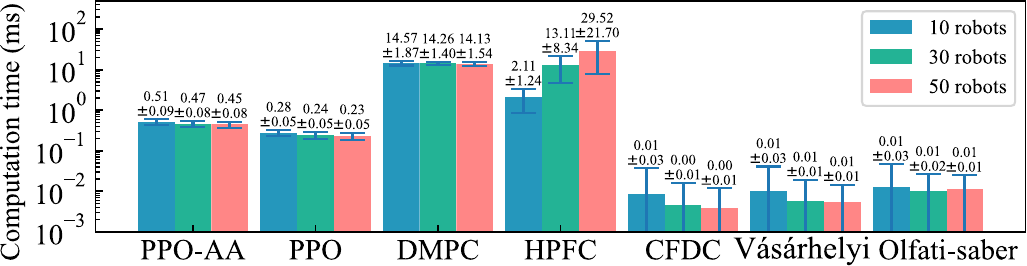}
    \caption{Computation time for each robot at different flocking scales.}
\label{fig:computation}
\end{figure}

The computation efficiency is evaluated based on the computation time of each algorithm in different environments. The computation costs of each algorithm for a different number of robots are shown in Fig. \ref{fig:computation}. All algorithms are tested on the same computer with Intel Core i7-10700K CPU.  PPO-AA and PPO are implemented in Python with PyTorch. DMPC is implemented in Python with OSQP solver\cite{stellato_osqp_2020}. 
HPFC, CFDC, Vásárhelyi{'s}, and Olfati-saber{'s} algorithm are implemented in C++. Results indicate that learning-based methods are more efficient compared to optimization-based methods. Furthermore, they are only slightly slower than rule-based methods. The computation time remains consistent regardless of the number of robots, demonstrating the scalability of the learning-based approach.

\begin{figure*}[tbp]
    \centering
    \includegraphics[width=\linewidth]{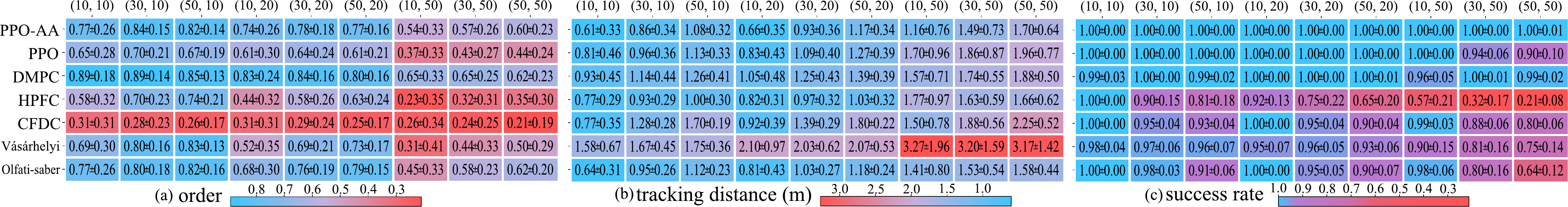}
    \caption{{Flocking order} $\Phi_{o}$ (a), {Tracking distance} $\Phi_{t}$ (b), and {Success rate} $ \Phi_{s}$ (c) at different environments with $(|\mathcal{V}|, |\mathcal{V}_o|)$, where $|\mathcal{V}|$ is the number of robots and $|\mathcal{V}_o|$ is the number of obstacles, $0\leq \Phi_{o}\leq 1$ and $0\leq \Phi_{s}\leq 1$. A larger value for $\Phi_{o}$ means better performance in terms of flocking order. A smaller $\Phi_{t}$ means better performance in terms of position tracking. A larger value for $\Phi_{s}$ means better performance in terms of motion safety. 
    }
    \label{fig:metrics}
\end{figure*}

\subsubsection{Flocking optimality}
The optimality of flocking control is evaluated using two metrics: \emph{flocking order} and \emph{tracking distance}. The two metrics are assessed only for robots that do not collide with each other or obstacles. The \emph{flocking order} indicates the consistency of robots \cite{soria_distributed_2022}, which is defined as
\begin{equation}
\label{eq:order}
    \Phi_{o} = \sum\nolimits_{i\in\mathcal{V}} \sum\nolimits_{j\in\mathcal{N}_i} \frac{\boldsymbol{v}_i^\top \boldsymbol{v}_j } {|\mathcal{V}||\mathcal{N}_i|\Vert \boldsymbol{v}_i \Vert \Vert \boldsymbol{v}_j \Vert}
\end{equation}
The \emph{tracking distance} metric measures the tracking performance of flocking control, which is given by 
\begin{equation}
    \label{eq:tracking_distance}
    \Phi_{t} = \left.\sum\nolimits_{i\in\mathcal{V}} \Vert \boldsymbol{p}_i - \boldsymbol{p}_c \Vert\right/|\mathcal{V}|
\end{equation}
All algorithms are evaluated in $9$ different environment settings with the robot number of $10$, $30$, or $50$ and the obstacle number of $10$, $30$, or $50$, respectively.

The proposed learning-based methods, PPO-AA and PPO, demonstrate high flocking order across various tasks as illustrated in Fig. \ref{fig:metrics}(a), highlighting their effectiveness in maintaining cohesive robot groups. In comparison, CFDC and Vásárhelyi{'s algorithm exhibit} a lower flocking order due to their control rules not accounting for oscillations. DMPC shows a slightly higher flocking order than the learning-based methods at the cost of tracking distance and motion safety, which will 
be verified later on.


PPO-AA achieves the shortest tracking distance among the evaluated methods as shown in Fig. \ref{fig:metrics}(b), demonstrating its superior performance in minimizing the distance between robots and the flocking {desired position}. In contrast, DMPC struggles with tracking distance due to the inherent difficulty in balancing multiple objectives. DMPC’s performance is dependent on carefully tuned parameters, which require extensive expert knowledge. While rule-based methods can achieve short tracking distances when parameters are well-tuned (\emph{e.g.} CFDC and Olfati-Saber{'s} algorithm), they may perform poorly as observed in the Vásárhelyi{'s} algorithm. Overall, the learning-based method PPO-AA stands out for its exceptional tracking distance performance.

In summary, it is difficult for rule-based methods to balance multiple objectives of flocking, while {optimization-based} methods have the potential but require 
extensive expert tuning efforts. In contrast, learning-based methods demonstrate competent performance across various metrics, showcasing their ability to achieve optimal results in diverse aspects. 



\subsubsection{Motion safety}
\label{sec:obstacle_avoidance_performance}
Motion safety is evaluated based on the success rate of collision avoidance for robots in flocking, whose metric is given by $\Phi_{s} = \left.{|\mathcal{V}_s|}\right/{|\mathcal{V}|}$, 
where $|\mathcal{V}_s|$ is the number of robots without collision. Both rule-based methods and DMPC would experience apparent performance degeneration in terms of motion safety with the complexity increase of the environment, for example, the increase of robot numbers and the increase of obstacle numbers, as shown in Fig. \ref{fig:metrics}(c). This is due to the fact that parameters tuned for a simple scenario are barely adaptable to a complex one for both rule-based methods and DMPC. The vanilla PPO method also {experiences} significant performance degeneration due to the lack of motion intention anticipation. In contrast, PPO-AA consistently exhibits the highest collision avoidance success rate in most cases, underscoring the safety of the learning-based distributed flocking control framework.




\subsection{Ablation study}
\begin{figure}[htbp]
\centering
    \includegraphics[width=1\linewidth]{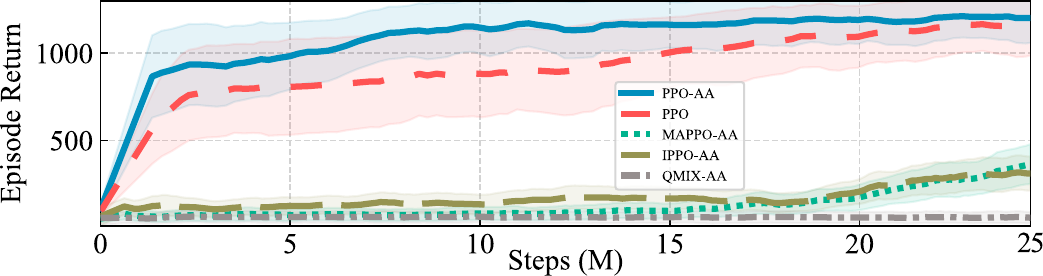}
    \caption{{Learning curves} of different learning algorithms. }
    \label{fig:training_curve}
\end{figure}

The effectiveness of the credit assignment mechanism (described in Sec. \ref{sec:gibbs_payoff_distribution}) and the action attention structure (detailed in Sec. \ref{sec:action_attention}) is evaluated through an ablation study. For the credit assignment mechanism, we compare the performance of several CTDE frameworks: Multi-agent PPO with action attention (MAPPO-AA) \cite{yu_surprising_2022}, independent PPO with action attention (IPPO-AA) \cite{yu_surprising_2022}, and QMIX with action attention (QMIX-AA) \cite{rashid_qmix_2018}, as shown in Fig. \ref{fig:training_curve}. The results reveal that CTDE frameworks (MAPPO, IPPO, QMIX) struggle to learn effective distributed policies in our 30-robot training scenario. However, the DTDE by credit assignment can further improve data efficiency.

For the action attention structure, the learning curves in Fig. \ref{fig:training_curve} demonstrate that PPO-AA outperforms PPO in terms of data efficiency and training variance. Also, PPO-AA exhibits higher optimality and safety, as illustrated in Fig. \ref{fig:metrics}. These results highlight the superior effectiveness of the action attention structure in enhancing both the performance and safety of the learning-based method.

\subsection{Generalization validation}
\begin{figure}[htbp]
    \centering
    \includegraphics[width=1\linewidth]{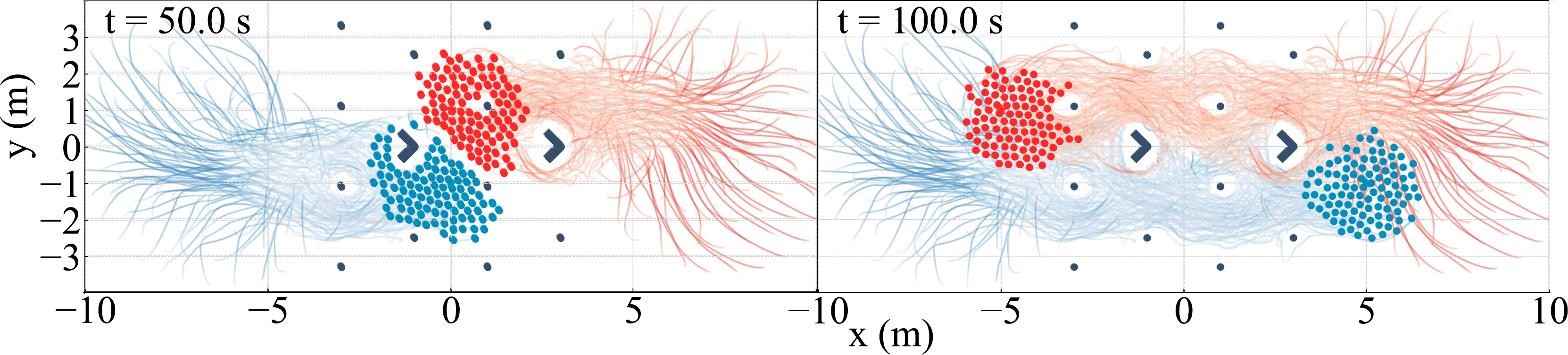}
    \caption{{Generalization validation.} Two flocks move in an opposite direction in an environment with  $22$ obstacles (including $2$ non-convex obstacles), each of which contains $100$ robots. No collision is observed in 
    this simulation.}
    \label{fig:large_scale_crossing}
\end{figure}

The algorithm generalization performance is further evaluated in 
a scenario involving two separate flocks moving in shared space as shown in Fig. \ref{fig:large_scale_crossing}. Each flock with $100$ robots moves at a speed of $0.1 \mathrm{m/s}$ in {the} opposite direction in an environment with $22$ obstacles (including $2$ non-convex obstacles). The results demonstrate that the learned flocking control by our method performs effectively in larger-scale and dynamically changing environments, highlighting its impressive generalization ability.

\subsection{Real-world experiment}
\begin{figure}[htbp]
    \centering
    \includegraphics[width=1\linewidth]{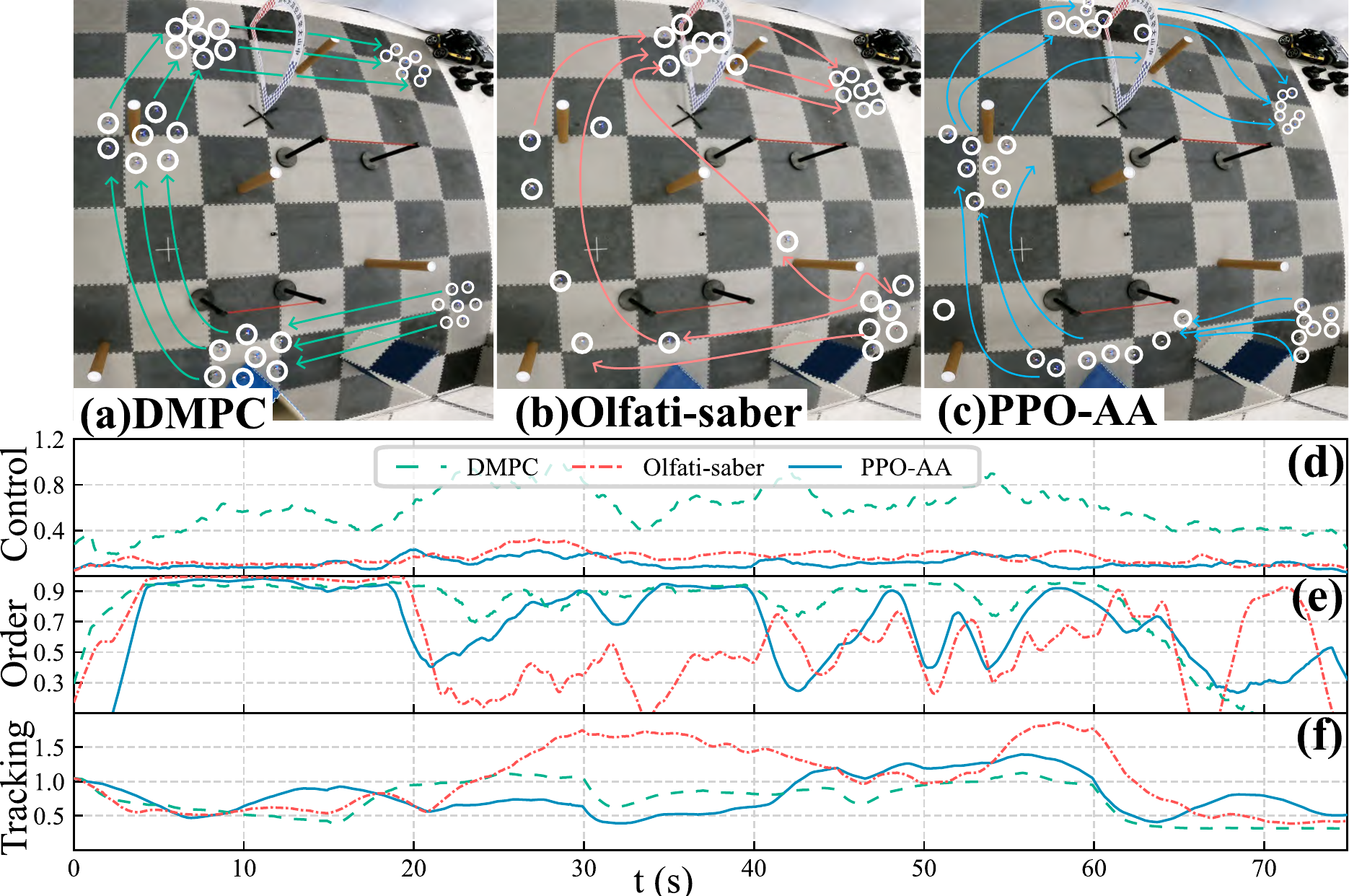}
    \caption{{Real-world experiments. Flocks with 7 robots are expected to move clockwise around a rectangular path. Metrics are shown in (d) control inputs ($\mathrm{m/s^2}$), (e) flocking order, and (f) tracking distance ($\mathrm{m}$).}}
    \label{fig:experiment_compare}
\end{figure}

To validate the algorithm performance in real-world conditions, we designed a complex environment featuring cylindrical obstacles, narrow passages, and a non-convex obstacle as shown in Fig. \ref{fig:experiment_compare}. The robot flocks are expected to move clockwise around a rectangular path within $1$ minute as illustrated in Fig. \ref{fig:experiment_compare}.
The experiments are conducted with Crazyflie drones and FZMotion\footnote{https://www.lusterinc.com/FZMotion/} motion capture system via Crazyswarm interface \cite{preiss_crazyswarm_2017}. The drone positions are tracked at $120\mathrm{Hz}$, while acceleration commands of flocking control algorithms are sent to drones at $10\mathrm{Hz}$. Onboard controller \cite{mellinger_minimum_2011} tracks the acceleration commands.

We compared the performance of our method (PPO-AA) with the two best benchmark algorithms in simulation, \emph{e.g.}, DMPC, and Olfati-saber{'s} algorithm. The experiments revealed that the Olfati-saber{'s} algorithm struggled to maintain the flock’s cohesion in densely obstructed environments, frequently losing track of the path. DMPC, on the other hand, maintained an optimal shape. The PPO-AA approach exhibited superior adaptability under uncertainty{, showing a success rate of about $99\%$}. When navigating narrow passages, the flock transformed into a linear formation, reducing the risk of collision. Once leaving the narrow passage, robots would recover a tight configuration, showcasing the robustness and collision avoidance capabilities of PPO-AA. 

We evaluated the motion performance of three algorithms in real-world experiments including control input, flocking order, and tracking distance. 
{As shown in Fig. \ref{fig:experiment_compare}, Olfati-saber's algorithm has the worst performance in flocking order and tracking distance. DMPC boasts the highest flocking order, yet PPO-AA closely trails behind, not far off in performance. Notably, PPO-AA's tracking distance is highly competitive with DMPC, showcasing the strides our method has made.}

Furthermore, we designed an experiment in which two flocks move in the same environment in an opposite direction and regard each other as dynamic obstacles, as shown in Fig. \ref{fig:top_figure}. The experiment demonstrates the ability of our method to avoid dynamic obstacles in a real-world implementation.

\section{Conclusion}
In this paper, we proposed a learning-based distributed flocking control framework based on a GRF. A credit assignment is introduced to achieve decentralized training and decentralized execution. Implicit intention anticipation is achieved via an action attention structure. Numerical simulation results demonstrated that our method is more computationally efficient than optimization-based methods. Our method shows better performance over both optimization-based approaches and rule-based solutions in terms of flocking performance and motion safety in various environments. The ablation study illustrated the effectiveness of the credit assignment and action attention structure.
Real-world experiments further demonstrated the competence of our method.

\ifCLASSOPTIONcaptionsoff
  \newpage
\fi



\bibliography{reference}
\bibliographystyle{IEEEtran}

\end{document}